\documentclass[journal]{IEEEtran}

\usepackage{soul} 

\usepackage{color} 

\usepackage{tabularx} 
\newcolumntype{Y}{>{\centering\arraybackslash}X} 

\usepackage{multirow} 

\newcolumntype{?}{!{\vrule width 1pt}}

\ifCLASSINFOpdf
   \usepackage[pdftex]{graphicx}
\else
\fi

\usepackage{amsmath}

\usepackage{tikz}
\usepackage{textcomp}
\usepackage{hyperref}
\usepackage{lipsum}

\newcommand\copyrighttext{%
	\footnotesize \textcopyright 2020 IEEE. Personal use of this material is permitted.
	Permission from IEEE must be obtained for all other uses, in any current or future 
	media. 
	DOI: \href{https://ieeexplore.ieee.org/document/9043749}{10.1109/TAFFC.2020.2981610}}
\newcommand\copyrightnotice{%
	\begin{tikzpicture}[remember picture,overlay]
	\node[anchor=south,yshift=10pt] at (current page.south) {\fbox{\parbox{\dimexpr\textwidth-\fboxsep-\fboxrule\relax}{\copyrighttext}}};
	\end{tikzpicture}%
}

\begin{document}

\title{A Bayesian Deep Learning Framework for End-To-End Prediction of Emotion from Heartbeat}
\author{Ross~Harper\textsuperscript{1},~\IEEEmembership{Member,~IEEE,}~and~Joshua~Southern\textsuperscript{1},~\IEEEmembership{Member,~IEEE}
\thanks{\textsuperscript{1}Ross Harper and Joshua Southern are with Limbic Ltd, 82 Rivington Street, London, EC2A 3AZ, UK.}
\thanks{Correspondence to Ross Harper ({ross}@limbic.ai)}
}

\markboth{IEEE Transactions ON affective Computing}%
{Shell \MakeLowercase{\textit{et al.}}: Bare Demo of IEEEtran.cls for IEEE Journals}
\maketitle
\thispagestyle{empty}

\begin{abstract}
	Automatic prediction of emotion promises to revolutionise human-computer interaction. Recent trends involve fusion of multiple data modalities $-$ audio, visual, and physiological $-$ to classify emotional state. However, in practice, collection of physiological data `in the wild' is currently limited to heartbeat time series of the kind generated by affordable wearable heart monitors. Furthermore, real-world applications of emotion prediction often require some measure of uncertainty over model output, in order to inform downstream decision-making. We present here an end-to-end deep learning model for classifying emotional valence from unimodal heartbeat time series. We further propose a Bayesian framework for modelling uncertainty over these valence predictions, and describe a probabilistic procedure for choosing to accept or reject model output according to the intended application. We benchmarked our framework against two established datasets and achieved peak classification accuracy of 90\%. These results lay the foundation for applications of affective computing in real-world domains such as healthcare, where a high premium is placed on non-invasive collection of data, and predictive certainty.         
\end{abstract}

\copyrightnotice


\section{Introduction}
	\IEEEPARstart{H}{umans} are social creatures that evolved to think and communicate using emotional information. Cognition and emotion are thus intrinsically linked. Indeed, emotion has been shown to impact attention \cite{Ohman2001, Bar-Haim2007, Brosch2011}, memory \cite{Dolcos2004, Phelps2004, Sharot2004, Phelps2008}, perception \cite{Phelps2006, Brosch2010}, and decision-making \cite{Spence1995, Bechara1997, Bechara2005}. Automated analysis of human emotion has correspondingly garnered significant interest across academia and industry in recent years.
	
	A wealth of research within the field of affective computing has focussed on the analysis of face, voice and text data \cite{Pantic2000, Hanjalic2005, ElKaliouby2005, Yang2008, Zeng2009, Schuller2010, Polzehl2011, Schuller2011}. However, comparatively few studies investigate the prediction of emotion from physiological signals. The link between emotion and physiology has neurobiological foundations in the limbic and autonomic nervous systems. The limbic system, which includes structures such as the amygdala and hippocampus, is important for the processing of emotional information \cite{Phillips2003, Dalgleish2004, Redondo2014}. Physiological responses to emotional stimuli are then coordinated by another limbic structure, the hypothalamus, which regulates heartbeat through antagonistic activity in the sympathetic and parasympathetic branches of the autonomic nervous system (ANS) \cite{Ekman1983, Kreibig2010}. 
	
	For ease of reference, we refer to audio-, visual-, and physiology-based methods of emotion detection as ED\textsubscript{A}, ED\textsubscript{V}, and ED\textsubscript{P} respectively. ED\textsubscript{P} has tremendous potential to compliment alternative methods of affective computation. ED\textsubscript{A} and ED\textsubscript{V} rely heavily on expression, which varies across individuals and cultures \cite{Ekman1987, Scherer2001} and leaves room for deception. By comparison, physiological processes are far less volitional. ED\textsubscript{P} further presents an opportunity for non-invasive continuous monitoring. Physiological signals may be passively analysed throughout the day, whereas audiovisual data is rarely available or feasible to collect continuously. ED\textsubscript{P} therefore has the potential to fill critical gaps in domains such as healthcare (where physiological data is already abundant), transport, and hospitality. 
	
	To date, there exist a range of affordable wearable monitoring devices that possess the capacity for high quality heartbeat recording \cite{Stahl2016, Wallen2016}. These devices have already been used to detect cardiac abnormalities such as atrial fibrillation \cite{Tison2018}. However, while heartbeat data is common, other physiological signals are strikingly rare in today's technology landscape. To be immediately relevant, therefore, ED\textsubscript{P} systems must be able to generate accurate predictions using only unimodal heartbeat input in the form of inter-beat-intervals (IBIs). Furthermore, the cardiac cycle is a complex dynamical process. The corresponding heartbeat time series is non-stationary \cite{Weber1992} and non-linear \cite{Sunagawa1998}. In order to adequately describe these characteristics, ED\textsubscript{P} systems must have the capacity to model complex temporal structure. 
	
	In many real-world scenarios, confidence is a key ingredient to decision-making (e.g. healthcare \cite{Johnson2016}). For applications in these domains, affective computing systems must have the capacity to describe uncertainty over their emotional state output. Consider a binary high/low valence classification problem. One might decide that valence predictions close to the class boundary (here, the valence scale mid-point) are more likely to be classified incorrectly. A solution could therefore be to only accept predictions that lie beyond a pre-specified distance from the class boundary (let's call this the `acceptance boundary'). The question remains $-$ how does one choose the acceptance boundary in a meaningful way? (Fig. \ref{fig:Framework}A). Furthermore, how do we translate this acceptance boundary to new environments where the data may have completely different statistical properties? Moreover, it is a strong assumption that predictions near to the class boundary are more likely to be incorrect (Fig. 1B). Indeed, a better approach would be to collect some measure of certainty in the model output. The decision on whether or not to accept the model's prediction can then be specified according to model confidence, rather than the output value. This matches the intuition that confidence is the key ingredient to decision-making, and permits simple translation of a given acceptance boundary between different datasets.

	In this study, we develop an end-to-end deep learning model for classifying emotional valence from unimodal IBI data. We implement recurrent and convolutional architectures to model temporal structure in the IBI signal, and further propose a Bayesian framework for modelling uncertainty in the output. We go on to describe a procedure for adjusting the acceptance boundary according to varying demands on confidence. This framework will be critical for applications of affective computing in domains such as healthcare, where a high premium is placed on predictive certainty and interpretability. We believe this is the first such model of its kind, has implications across the field of affective computing, and accelerates near-term relevance of ED\textsubscript{P} in real-world settings.

\section{Related Work}
	This section provides an overview of relevant work, with a focus on (A) analysis of unimodal heart data and temporal models for ED\textsubscript{P}, and (B) Bayesian neural networks.

	\subsection{Analysis of Unimodal Heart Data and Temporal Modelling} \label{section:Cardiocentric_and_temporal_models}
		Peripheral physiological markers of autonomic nervous activity include galvanic skin response (GSR), electroencephalogram (EEG), electromyogram (EMG), respiration, skin temperature (ST), electrocardiogram (ECG) and photoplethysmogram (PPG) \cite{Jerritta2011}. Note that a heartbeat time series can easily be extracted from both ECG and PPG in the form of inter-beat-intervals (IBIs). 
		
		Existing approaches for ED\textsubscript{P} typically pool a number of biosignals to form a multimodal input to some classifier algorithm \cite{Kim2008b,Alzoubi2012, Goshvarpour2017}. However, this multimodal approach contradicts the vast amount of physiological data obtained by affordable wearable devices today (which is almost exclusively unimodal IBI data). Comparatively few studies narrow their scope in accordance with these practical limitations. 

		Those studies that have explored unimodal IBI models for emotion detection tend to ignore temporal structure of the signal. Instead, they employ `static' classifiers that process global features from the input time series (or features calculated over a number of shorter segments). Such approaches include Naive Bayes (NB) \cite{Miranda-Correa2017b, Subramanian2016}, linear discriminant analysis (LDA) \cite{Agrafioti2012}, and support vector machine (SVM) \cite{Katsigiannis2018, Guo2016a, Valenza2014b}. A summary can be found in Table \ref{relevant_work}.
		
		A number of studies have sought to model temporal information within EEG signals, using hidden Markov models \cite{Torres-Valencia2015}, Gaussian Process models \cite{Garcia2016}, continuous conditional random fields \cite{Soleymani2012}, and long short-term memory (LSTM) neural networks \cite{Alhagry2017}. Such temporal treatment, however, is rare for other peripheral physiological signals. One notable exception to this involved the use of a temporal neural network to predict valence from ECG input \cite{Keren2017}. Here, a combination of convolutional and recurrent layers performed end-to-end learning, improving on computationally expensive manual feature engineering schemes. In this study, we too implement end-to-end learning. However, in accordance with aforementioned practical limitations, we further limit model input to IBI time series in order to simulate data of the form generated by consumer wearables.
		
		Convention within affective computing is to implement an experimental protocol where an emotion-inducing stimulus is combined with participant self-reporting of affective state. However, there is a large amount of variability between self-reporting frameworks. A clear division concerns discrete and dimensional theories of emotion \cite{Barrett1998}. To construct our Bayesian framework, we require only a clear measure affective state that can be easily compared with existing benchmarks. For the purposes of this study, we choose to use emotional valence, which is suited to both a classification and regression setting, and avoids issues of semantic confusion that can accompany the terminology around emotional/physiological arousal in the cardiovascular literature.
		
		At this point, we wish to point out a stark absence of consensus within the literature around data subsetting for machine learning. A typical experimental setup can yield multiple input-output pairs for a single study participant. Many studies partition train, validation, and test datasets without reference to the study participant from which the data was generated. However, ECG has been shown to exhibit subject-specificity \cite{Kolodyazhniy2011}. It is therefore  unsuitable to include data from a given participant in both the train and test/validation subsets. Moreover, real-world applications may not permit subject-specific calibration, making models that can generalise to new individuals increasingly useful. For these reasons, we propose a sensible evaluation method to be leave-k-subjects-out (LkSO) cross-validation, which has been used previously \cite{Miranda-Correa2017b, Ferdinando2016, Agrafioti2012} and will be adopted in this study.   
		
		\begin{table*}[!t]
			\renewcommand{\arraystretch}{1.3}
			\caption{Summary of relevant work - classification of emotion from heartbeat}
			\label{relevant_work}
			\centering
			\begin{tabularx}{\linewidth}{|Y|Y|c|c|Y|Y|Y|}
				\hline
				\hline
				Author & Stimulus & Subjects & LkSO & Model & Target & Performance\\
				\hline
				Katsigiannis \& Ramzan 2018 & Videos & 23 & No & SVM & High/Low Valence & F1. 0.5305 (Chance: 0.500) \\
				\hline
				Subramanian et al 2018 & Videos & 58 & No & NB & High/Low Valence & Acc. 60\% (Chance: 50\%) \\
				\hline
				Miranda-Correa et al 2017 & Videos & 40 & Yes & NB & High/Low Valence & F1. 0.545 (Chance: 0.500) \\
				\hline
				Guo et al 2016 & Videos & 25 & No & SVM & High/Low Valence & Acc. 71.40 (Chance: 50\%) \\
				\hline
				Ferdinando et al 2016 & Videos \& IAPS & 27 & Yes & KNN & High/Medium/High Valence & Acc. 59.2\% (Chance: 33.3\%) \\
				\hline
				Valenza et al 2014 & IAPS & 30 & No & SVM & High/Low Valence & Acc. 79.15\% (Chance: 50\%) \\
				\hline
				Agrafioti et al 2012 & IAPS & 32 & Yes & LDA & Gore, Erotica & Acc. 46.56\% (Chance: 50\%) \\
				\hline
			\end{tabularx}
		\end{table*}
         
	\subsection{Bayesian Neural Networks}
	Despite the widespread success of deep learning, traditional neural networks lack probabilistic considerations. This results in the model outputting a single point estimate for each input sample, which can lead to overconfidence in downstream decision-making. This is a serious issue for applications where datasets are imbalanced and representing uncertainty is of critical importance (e.g. medical diagnosis) \cite{Ghahramani2015}. 
	
	To combat this, neural networks may be re-cast as Bayesian models in order to capture probability in the output. In this formalism, the weights of the network, $\boldsymbol{\theta}$, are said to belong to some prior distribution, $p(\boldsymbol{\theta})$. A posterior distribution over the network weights can then be conditioned on the data, $\boldsymbol{D}$, according to Bayes' rule:
	
	\begin{equation} \label{eq1}
		\begin{split}
			p(\boldsymbol{\theta}|\boldsymbol{D}) = \frac{p(\boldsymbol{D}|\boldsymbol{\theta})p(\boldsymbol{\theta})}{p(\boldsymbol{D})}
		\end{split}
	\end{equation}
	
	where $p(\boldsymbol{\theta}|\boldsymbol{D})$ is the posterior distribution of the network weights and  $\boldsymbol{D}$ are the training data, which for a supervised task comprises the input-output sample pairs (in our case, the input is the IBI time series and the output is the subject self-reported valence). 
	
	When making a prediction on a new input sample, $\mathbf{x}$, we can measure our uncertainty in the output, $y$, using the posterior predictive distribution, $p(y|\mathbf{x}, \boldsymbol{D})$, which is calculated by marginalising out the network parameters:
	
	\begin{equation} \label{eqnew}
		\begin{split}
			p(y|\mathbf{x},\boldsymbol{D}) = 
			\int p(y|\mathbf{x},\boldsymbol{\theta})p(\boldsymbol{\theta}|\boldsymbol{D})d \boldsymbol{\theta}
		\end{split}
	\end{equation}
	
	While useful from a theoretical perspective, such Bayesian treatment of neural networks is infeasible to compute. Indeed, the evidence term in the denominator of Equation \ref{eq1} (also known as the marginal probability) amounts to the integral over all possible values of the network weights:
	
	\begin{equation} \label{eq2}
		\begin{split}
			p(\boldsymbol{D}) = \int p(\boldsymbol{D}|\boldsymbol{\theta})p(\boldsymbol{\theta})d\boldsymbol{\theta}
		\end{split}
	\end{equation} 
	
	For obvious reasons, exact posterior inference is rarely achievable. Instead, we seek to approximate these posterior distributions. Early attempts at this include Monte Carlo (MC) \cite{Neal1996} or Laplace \cite{MacKay1992} approximation methods. However, these are slow and computationally expensive when applied to modern deep learning architectures. Research in the field has focussed on identifying faster inference methods such as stochastic gradient Langevin diffusion \cite{Welling2011}, expectation propagation \cite{Hasenclever2015}, and variational methods \cite{Graves2012}. 
	
	Interestingly, Bayesian neural networks can also be constructed using Monte Carlo dropout - a widely-used approach to reduce over-fitting during network training \cite{Gal2016}. Dropout is a process by which individual nodes within the network are randomly removed during training according to a specified probability \cite{Sutskever2014}. By implementing dropout at \textit{test} (as well as in training) and performing $N$ stochastic forward passes through the network, we can approximate a posterior distribution over model predictions (approaching the true distribution as $N\to \infty$ ). In this paper, we implement Monte Carlo dropout as an efficient way to describe uncertainty over emotional state predictions.

\section{Model Architecture} \label{section:ProposedFramework}
	An overview of our model architecture is shown in Fig. \ref{fig:ModelArchitecture}. Data flows through two concurrent streams. One stream comprises four stacked convolutional layers that extract local patterns along the length of the time series. Each convolutional layer is followed by dropout and a ReLU activation function. A global average pooling layer is then applied to reduce the number of parameters in the model and decrease over-fitting. The second stream comprises a bidirectional LSTM followed by dropout. This models both past and future sequence structure in the input. The output of both streams is then concatenated before passing through a dense layer to output a regression estimate for valence.
	
\section{Bayesian Framework} \label{section:ProposedFramework2}

	In order to capture uncertainty in model predictions, dropout is applied at test time. For a single input sample, stochastic forward propagation is run $N$ times to generate a distribution over model output. This empirical distribution approximates the posterior probability over valence, given the input IBI time series. 
	
\begin{figure}
		\centering
		\includegraphics[width=\linewidth]{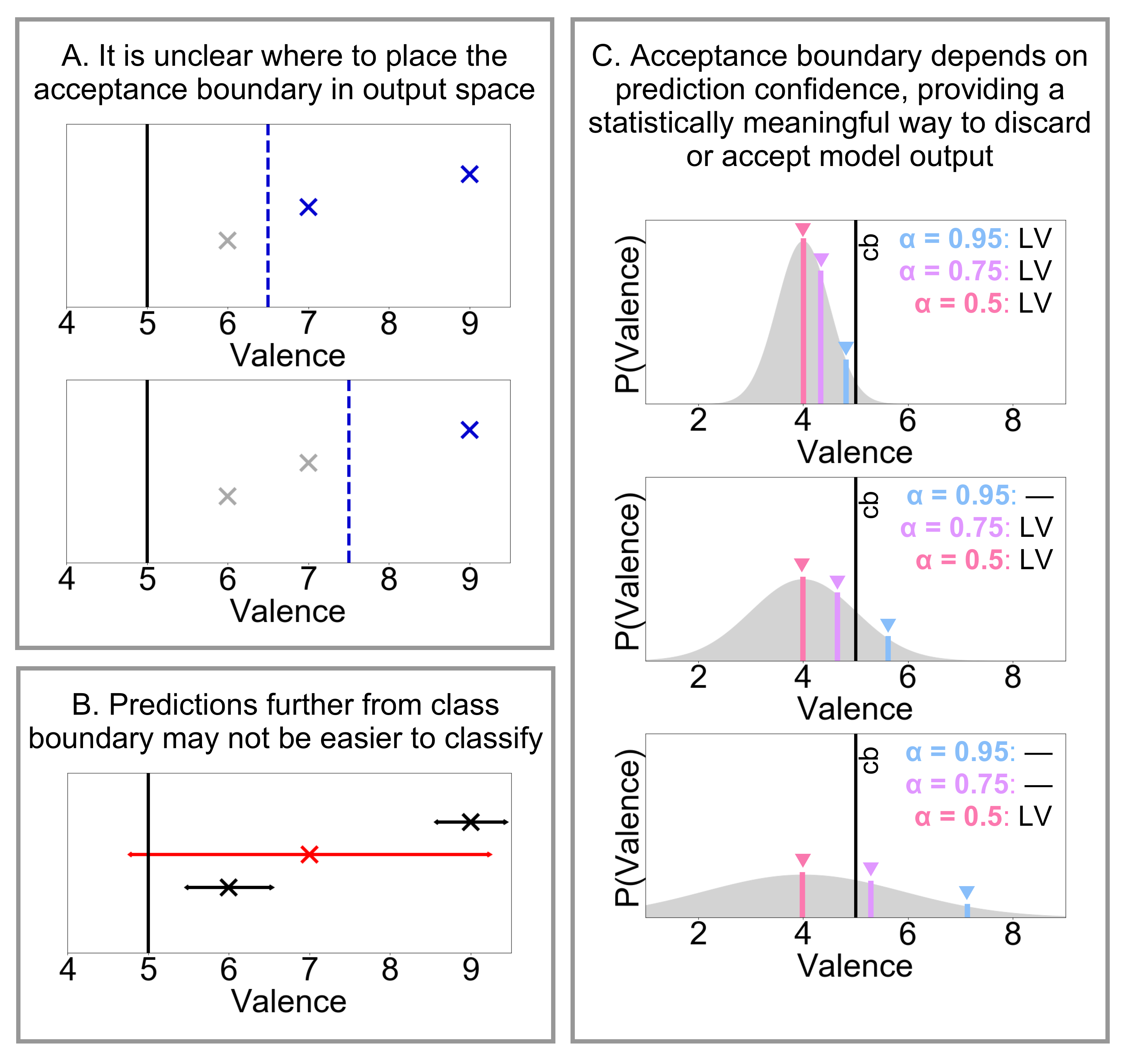}
		\caption{The need for probabilistic frameworks in affective computing. (A) One might introduce an acceptance boundary (`ab', blue dashed line) at a pre-specified distance from the class boundary (`cb', black line). However, it is unclear where to place this acceptance boundary (compare top and bottom), which has a large impact on what samples get classified (blue). (B) It is not necessarily true that predictions further from the class boundary are more likely to be accurate. We see that a prediction further away from the class boundary has a greater level of associated error (red) compared to other model predictions (grey). Here, it is prediction confidence that is critical, not absolute value. (C) Our Bayesian framework illustrated for a binary classification problem. A probabilistic model outputs the posterior probability over valence. These predictions are either accepted or rejected according to a confidence threshold, $\alpha$ (illustrated for $\alpha = 0.95, 0.75,$ and 0.5 on three example output distributions, which have the same mode but vary in certainty). The model either classifies `low valence' (LV) or `I don't know', depending on how confident we wish our model to be.}
		\label{fig:Framework}
	\end{figure}

\begin{figure}
		\centering
		\includegraphics[width=\linewidth]{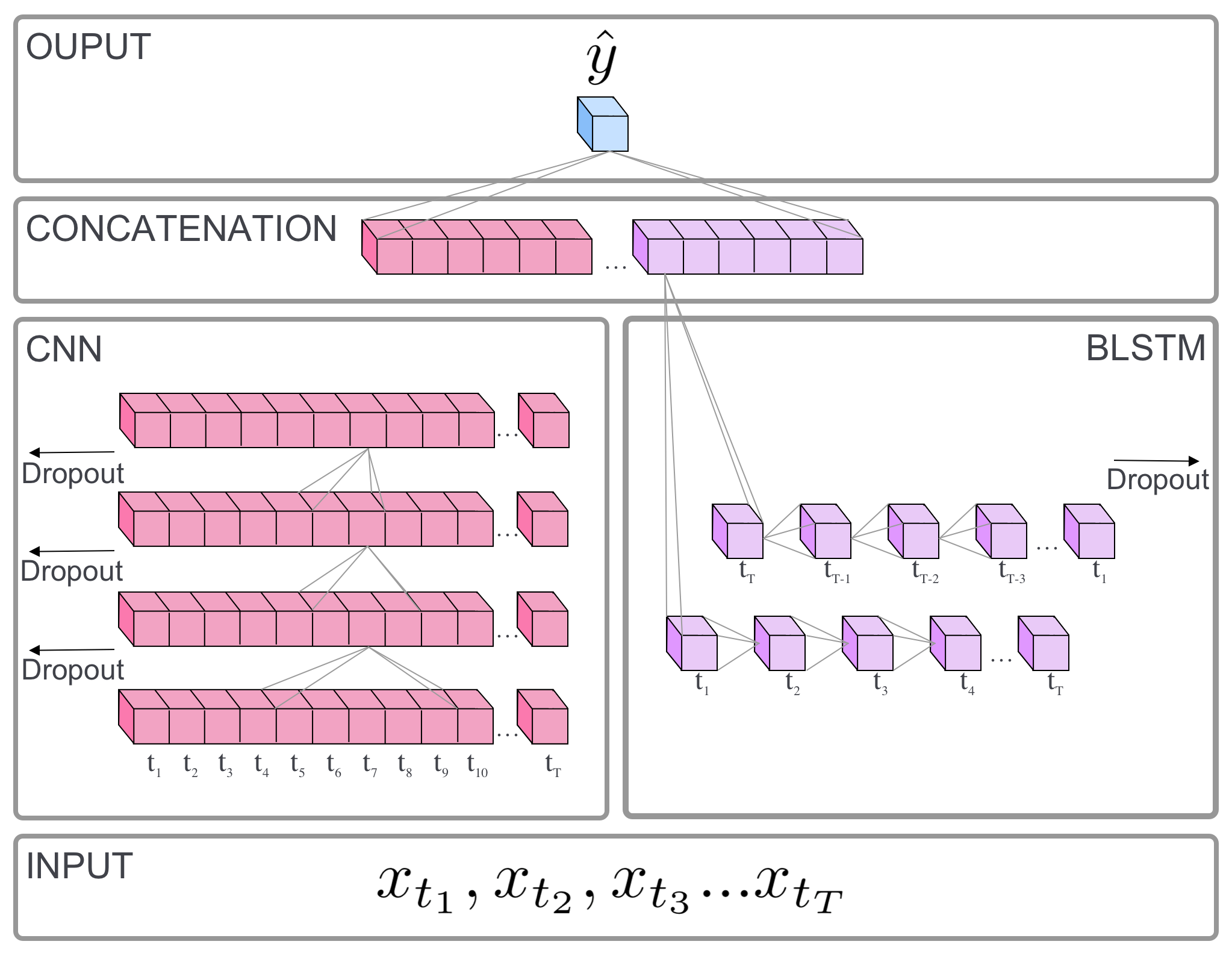}
		\caption{End-to-end model architecture. IBI time series data ($x_{t_1}, x_{t_2}, ... x_{t_T}$) flows through two temporal processing streams: 1D convolutions (pink) and a bi-directional LSTM (purple). The output from both streams is then concatenated before passing through a dense layer to output a regression estimate for valence, $\hat{y}$. }
		\label{fig:ModelArchitecture}
		\end{figure}

	For regression problems, the reader may stop here. In order to translate from a regression to a classification scheme (as was used in \cite{Miranda-Correa2017b} and \cite{Katsigiannis2018}, and will be implemented in this study for comparison), we introduce class boundaries in continuous space. For a binary classification scheme, the class boundary is naturally along the central point of the valence scale to delimit two class zones (high and low valence). We next introduce a confidence threshold, $\alpha$, which acts as the acceptance boundary and determines whether a prediction is accepted or rejected according to how confident we wish to be. For example, when $\alpha = 0.95$, we specify that at least $95\%$ of the output distribution mass must lie within a given class zone in order for the input sample to be classified as belonging to that class (Fig. \ref{fig:Framework}C). If this is not the case, no prediction is made (i.e. the model answers, `I don't know'). As our model may not classify all instances, we also adopt the term `coverage' to denote the proportion of samples for which it is confident enough to make a prediction.
	
	Note that for a binary classification problem, and where $N$ is an odd number, there will always be at least 50\% of the output distribution mass within one of the two class zones. Thus, when $\alpha = 0.5$, classification is determined by the median of the output distribution (Fig. \ref{fig:Framework}), and the coverage is 100\%. As $\alpha$ increases, model behaviour moves from risky to cautious $-$ lower coverage, but predictions are more certain. 
	
\section{Data}
	We applied our Bayesian deep learning framework for end-to-end prediction of emotion using heartbeat (IBI) data extracted from two established datasets $-$ AMIGOS \cite{Miranda-Correa2017b} and DREAMER \cite{Katsigiannis2018}. In this section, we provide details on these data, which were chosen for their quality, clarity, and comparability.
	
	\subsection{AMIGOS}
		These data include 40 healthy participants (13 female; 27 male) aged between 21 and 40 years old (mean: 28.3). ECG data was recorded using a Shimmer\textsuperscript{TM} ECG wireless monitoring device (256 Hz, 12 bit resolution). Subjects watched 16 short videos (duration $<250$ seconds) that had been previously scored for emotional content. The videos were presented in a random order with each trial comprising a 5-second baseline recording showing a fixation cross, presentation of the video stimulus, followed by self-assessment of valence on a scale of 1 to 9 using the self-assessment manikin (SAM) \cite{Morris1995}. 
		
	\subsection{DREAMER}
		These data include 25 healthy participants (11 female; 14 male) aged between 22 and 33 years old (mean: 26.6). ECG data was also recorded using a Shimmer\textsuperscript{TM} ECG wireless monitoring device (256 Hz, 12 bit resolution). Subjects watched 18 short film clips (duration: $<395$ seconds), which had been previously scored for their ability to elicit emotional responses \cite{Gabert-Quillen2015}. Each film clip was followed by self-assessment of valence on a scale of 1 to 5 using SAM \cite{Morris1995}, and was preceded by a neutral video presentation to establish baseline emotional state \cite{Gabert-Quillen2015}.

\section{Methods}	
	\subsection{Pre-processing}		
		To obtain data of the form generated by consumer wearables, IBIs were extracted from the ECG time-series using a combined adaptive threshold approach \cite{Christov2004}. This markedly reduced the information content of the input signal, making our task more challenging than described in \cite{Keren2017}. Nevertheless, inter-beat dynamics have previously been shown suitable for emotional state classification \cite{Valenza2014b}. The IBI time series were z-score normalised and zero padded to the length of the longest training sample. 
		
	\subsection{Training and Hyperparameters}
		Parameter search was used to select model hyperparameters. For this, a LkSO validation set of 4 subjects (k=4) was used to assess best-performance (lowest mean-squared loss) for a given combination of hyperparameters. Convolution kernels were initialised as He normal \cite{He2015} with filter size set to 128, and window size decreasing from 8 to 2 time steps with network depth. 50\% dropout was applied after each convolutional block, and 80\% dropout followed the bi-directional LSTM comprising 32 hidden units. Training was run for 1500 epochs using Adam optimisation \cite{Kingma2014}. Learning rate decreased from $e^{-3}$ to $e^{-4}$, halving with a patience of 100 epochs. Final model parameters were set to those associated with the lowest mean-squared loss on the validation set during training. The model was implemented using Tensorflow \cite{GoogleResearch2015}.
		
	\subsection{Evaluation}
		As discussed in Section \ref{section:Cardiocentric_and_temporal_models}, model performance was assessed using 10-fold leave-one-subject-out cross validation in order to generalise to new participants. Dropout was applied at test time with $N=1001$ forward propagations made through the network to generate an empirical distribution over model output. In accordance with the original studies from which we obtained our data, labels for valence were divided into high and low classes using the midpoint value of the SAM scale (5 for AMIGOS; 3 for DREAMER). As outlined in Section \ref{section:ProposedFramework}, a given test input sample was classified as high/low valence provided a proportion of at least $\alpha$ posterior distribution mass fell within a given class zone. If this was not the case, no prediction was made. We demonstrate the effect of adjusting $\alpha$ in Section \ref{section:Results}. Model accuracy was then calculated as the fraction of correct classifications over total predictions covered by the model.
		
\section{Results} \label{section:Results}
	 	
	 	\begin{table*}[ht]
	 		\caption{Comparison  of mean accuracy and F1 scores.  }
	 		\begin{center}
	 			\begin{tabularx}{\linewidth}{|c|c?c|c?Y|Y|Y|Y|Y|}
	 				\cline{3-9}
	 				\multicolumn{2}{c?}{}& \cite{Miranda-Correa2017b} & \cite{Katsigiannis2018} & LSTM (Non-Bayes) & CNN (Non-Bayes) & LSTM+CNN (Non-Bayes) & LSTM+CNN (Bayes $\alpha =$ 0.5) & LSTM+CNN (Bayes $\alpha =$ 0.9) \\
	 				\hline
	 				\multirow{2}{*}{AMIGOS} & Acc. & 0.54 & - & 0.47 &  0.69 & 0.79 & 0.81 & \textbf{0.90} \\
	 				& F1 & - & - & 0.49 & 0.65 & 0.78 & 0.80 & \textbf{0.88} \\
	 				\hline
	 				\multirow{2}{*}{DREAMER} & Acc. & - & 0.62 & 0.61 & 0.61 & 0.70 & 0.71 & \textbf{0.86} \\
	 				& F1 & - & 0.53 & 0.52 & 0.54 & 0.66 & 0.66 & \textbf{0.83}  \\
	 				\hline
	 			\end{tabularx}
	 		\end{center}
	 		\label{tab:comparison_performance}
	 	\end{table*}
	 	
	 		\begin{figure}
	 		\centering
	 		\includegraphics[width=\linewidth]{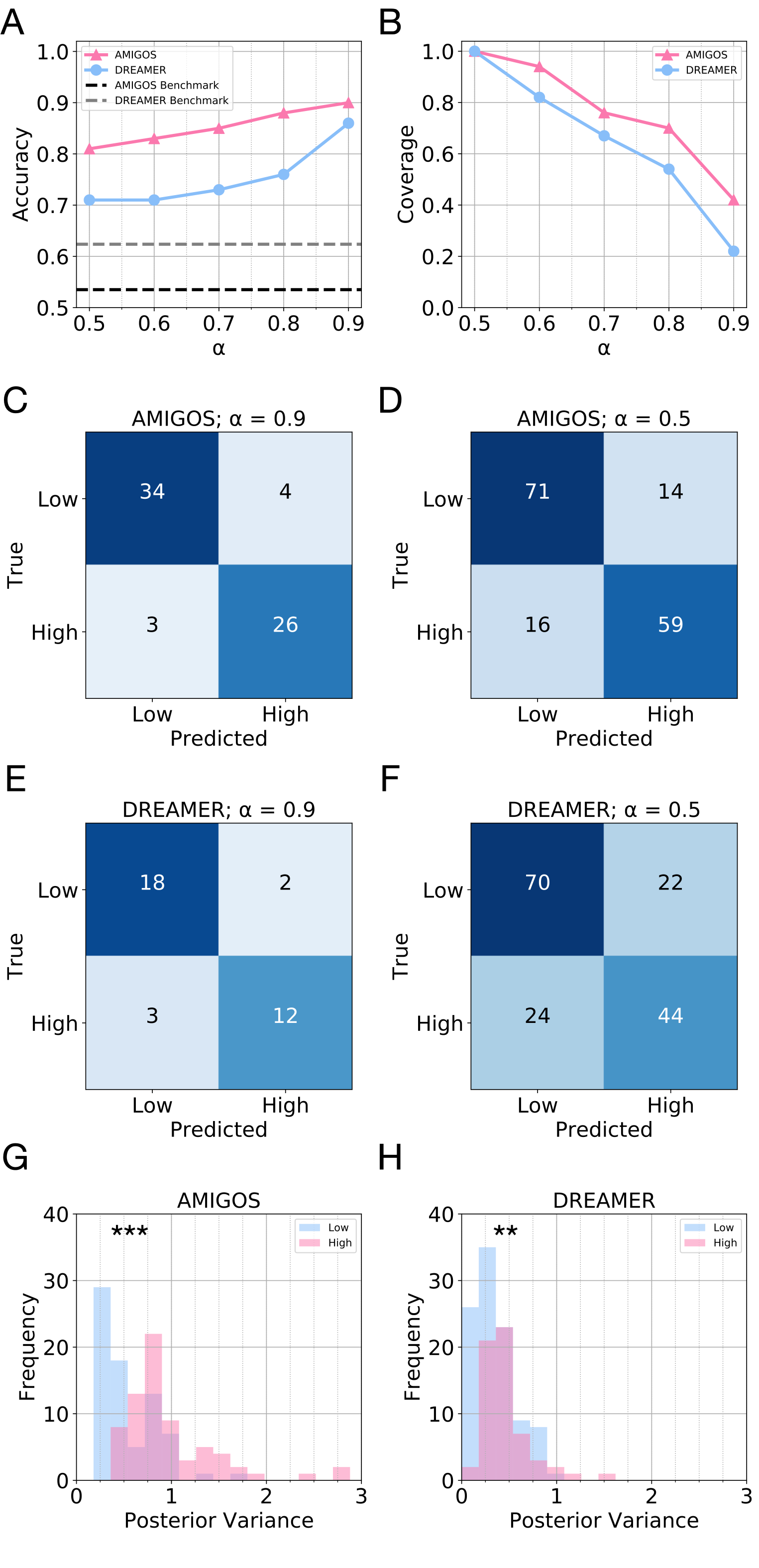}
	 		\caption{Results. (A) Model accuracy as a function of $\alpha$. Results are shown for AMIGOS (pink, triangles) and DREAMER (blue, circles) with benchmarks achieved by these original publications shown (black and grey dashed lines respectively). (B) Model coverage as a function of $\alpha$. Results are shown for AMIGOS (pink, triangles) and DREAMER (blue, circles). (C,D,E,F) Confusion matrices shown for AMIGOS (C,D) and DREAMER (E,F), with $\alpha$ set to 0.9 (C,E) or 0.5 (D,F). (G,H) Model uncertainty (as measured by variance of posterior output distribution) for low (blue) and high (pink) valence class labels. Significant differences observed for both AMIGOS (G, $p<1\times 10^{-11}$) and DREAMER (H, $p<0.001$)}
	 		\label{fig:ModelResults}
	 	\end{figure}
	 	
	 	To identify the benefit conferred by our temporal network architecture, we first evaluated our model without dropout at test time. In this non-Bayesian setting, model output was a single point estimate, which fell either in the high or low class zones, and was classified accordingly. Here, we achieved higher accuracy across both datasets than previously reported \cite{Miranda-Correa2017b, Katsigiannis2018} (Table \ref{tab:comparison_performance}). To further assess the effect of each network stream, we evaluated the CNN and LSTM stream separately (Table \ref{tab:comparison_performance}). The combined LSTM+CNN consistently outperformed either stream alone, indicating that the two streams each extract unique information relevant to the classification task.
	 	
	 	We next implemented our Bayesian framework with confidence threshold set to 50\% ($\alpha = 0.5$). As expected, maximal model coverage was observed (Fig. \ref{fig:ModelResults}B). Furthermore, no significant difference was found between the non-Bayes model and the Bayesian model when $\alpha = 0.5$ (Mann-Whitney U test; $p > 0.1$), demonstrating that our Bayesian framework does not lead to any performance decrease when coverage is 100\% (Table. \ref{tab:comparison_performance}). For clarity, recall that $\alpha$ denotes our confidence threshold on the model output, and is not used to denote the significance level of hypothesis tests.
	 	
	 	As the certainty threshold $\alpha$ increases, so too does classification accuracy, demonstrating a clear relationship between model confidence and propensity to make accurate predictions (Fig. \ref{fig:ModelResults}A, and Table \ref{tab:comparison_performance}). Naturally, as $\alpha$ increases, model coverage decreases due to the fact that fewer output distributions meet the necessary threshold for a prediction to be made. We see that with a 90\% confidence threshold ($\alpha = 0.9$), our model achieved peak accuracy for both datasets (Fig. \ref{fig:ModelResults}A, and Table \ref{tab:comparison_performance}). Interestingly, we found that certainty over model output was significantly greater for input time series that belong to the low valence class, for both datasets, as shown by Mann–Whitney U test (Fig. \ref{fig:ModelResults} G,H). This pattern is also reflected in the consistently better performance observed for the low valence class (Fig. \ref{fig:ModelResults} C,D,E,F).

\begin{figure}
	 		\centering
	 		\includegraphics[width=\linewidth]{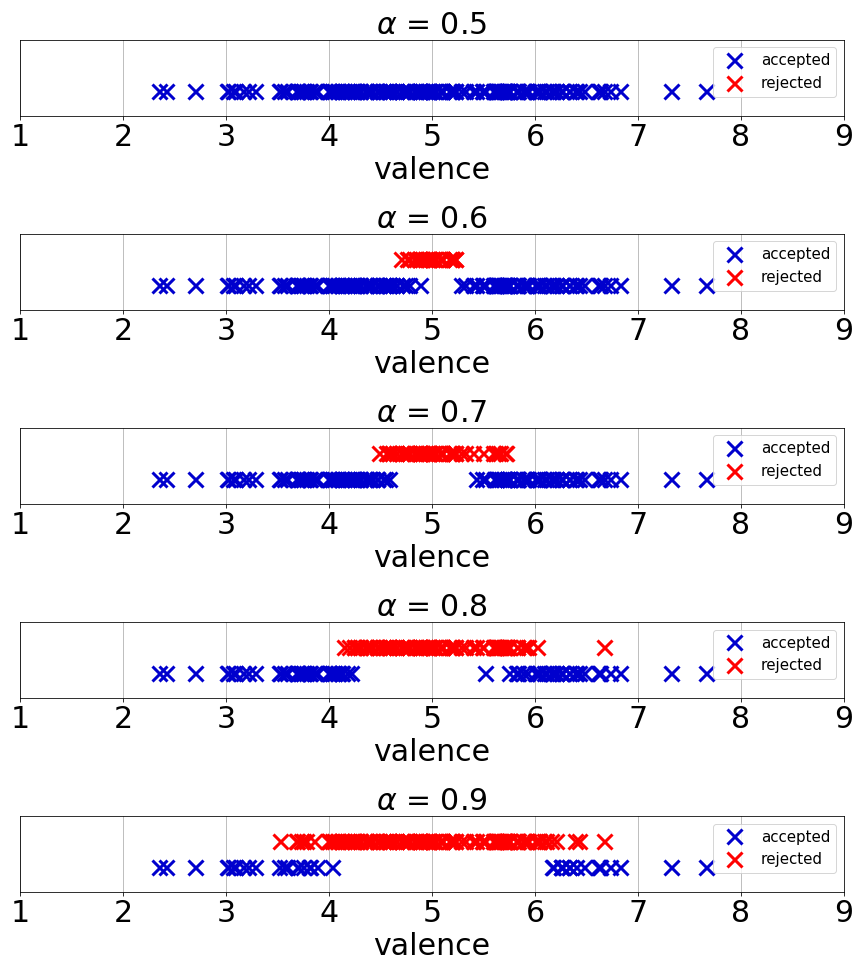}
	 		\caption{Acceptance or rejection of model predictions based on varying confidence thresholds. As confidence threshold ($\alpha$) increases from 0.5 to 0.9, a greater proportion of model predictions are rejected. Importantly, accepted (blue) and rejected (red) groups are not linearly separable in the output (valence) space, highlighting that the accept/reject decision is not equivalent to simply rejecting those predictions close to the class boundary (the valence midpoint). Our Bayesian framework matches the intuition that confidence is the key ingredient to decision-making (as illustrated in Fig. \ref{fig:Framework}).}
	 		\label{fig:AcceptanceBoundaries}
	 	\end{figure}

	 	We went on to plot valence predictions according to whether the prediction was accepted or rejected by our Bayesian framework. In Fig. \ref{fig:AcceptanceBoundaries}, we see that the two groups are not linearly separable in the valence space, highlighting that our acceptance boundary $-$ which is based on model certainty $-$ is not equivalent to simply ignoring model predictions close to the class boundary. The decision to reject or accept is a fundamental property of the model confidence in its own output.
	 	
\section{Discussion}

		The growing prevalence of high-fidelity, affordable wearable monitoring devices creates an opportunity for continuous emotion detection `in the wild'. The vast majority of existing approaches in the literature have relied on the fusion of multiple physiological signals for physiology-based emotion detection (or ED\textsubscript{P}). Although this multimodal treatment provides significant performance benefits, it is limited in practice. Indeed, the vast amount of physiological data obtained by affordable wearable devices today is almost exclusively unimodal IBI time series. Timely application of ED\textsubscript{P} in real world settings therefore requires models that are compatible with these input signals. 
		
		It has been shown previously that IBI extracted from PPG corresponds closely with IBI extracted from ECG \cite{Pietila2017, Harju2018}. This allowed us to exploit existing high-quality ECG datasets for the purposes of this study. We developed an end-to-end neural network to model temporal structure in the IBI time series and classify emotional valence. This outperformed previous benchmarks on these datasets \cite{Miranda-Correa2017b, Katsigiannis2018} and accelerates near-term relevance of ED\textsubscript{P} in real-world settings.
		
		Using Monte Carlo dropout, we re-cast our model as a probabilistic neural network in order to approximate the posterior predictive distribution over the output. This provides a crucial measure of uncertainty in the model predictions, which we used to specify a statistically meaningful acceptance boundary. By tuning the confidence threshold, $\alpha$, our Bayesian framework is able to trade off performance against coverage ($\alpha$ is itself a hyper-parameter for future cross-validation). In this study, we report peak model accuracy of 90\%. Further flexibility was achieved by framing our model as a regression problem, which allows the experimenter to specify class boundaries appropriate for binary- or multi-class tasks.
		
		Incorporating Bayesian considerations will drastically improve the applicability of affective computing in environments where confidence is a key ingredient to decision-making. Of particular interest to these authors is the application of emotion prediction to produce novel metrics within mental healthcare. For example, diagnosing the onset of major depressive disorder based on persistently low valence would reasonably require high levels of certainty. Additionally, clinical triaging systems based on predicted affective states may send uncertain predictions to a human expert for review. Similar levels of certainty may not, however, be necessary in consumer products. For example, content recommendations based on user mood might tolerate lower accuracy to ensure a greater number of recommendations. Our Bayesian framework provides a method of adjusting the acceptance boundaries in these different applications in a statistically meaningful way.
		
		Furthermore, generating a probability distribution over model outputs could provide a sensible way to combine spatiotemporal and multimodal data. Indeed, these signals could be analysed by separate neural networks, and the resulting predictions combined via Bayesian integration (which would necessarily require certainty measures of the kind generated by our Bayesian framework).
		
		Finally, our probabilistic framework provides deeper insight into the underlying properties of the data analysed. We found our model was more certain when classifying the low valence class for both datasets. This could be attributed to the subjective certainty of participants during their own self-reports. Alternatively, signatures of low valence within the heartbeat signal might contain more information than their high valence counterparts, requiring larger volumes of training data to reduce the uncertainty in the posterior predictive distribution. We look forward to future application of our Bayesian framework, and end-to-end deep learning model, to studies of emotional arousal, discrete models of emotion, multimodal input and indeed multimodal output.

\section{Conclusions}
	In this study, we developed an end-to-end deep learning model for classifying emotional valence from unimodal heartbeat data. Our temporal neural network architecture outperformed previous models on the AMIGOS \cite{Miranda-Correa2017b} and DREAMER \cite{Katsigiannis2018} datasets. We further proposed a Bayesian framework for modelling uncertainty over emotional state predictions, providing a means to tune prediction acceptance boundaries in a statistically meaningful way according to the confidence requirements for different tasks. That model performance improved with increasing certainty thresholds ($\alpha$) illustrates the potential for probabilistic modelling to enhance affective computing systems. Taken together, the component parts of this study represent an important step towards application of emotion detection in real-world settings, and provide a probabilistic standard for future work.
	
%

\ifCLASSOPTIONcaptionsoff
  \newpage
\fi

\bibliographystyle{ieeetr}
\bibliography{Bibliography}

\begin{thebibliography}{10}

\bibitem{Ohman2001}
A.~{\"{O}}hman, D.~Lundqvist, and F.~Esteves, ``{The face in the crowd
  revisited: A threat advantage with schematic stimuli},'' {\em Journal of
  Personality and Social Psychology}, vol.~80, no.~3, pp.~381--396, 2001.

\bibitem{Bar-Haim2007}
Y.~Bar-Haim, D.~Lamy, L.~Pergamin, M.~J. Bakermans-Kranenburg, and M.~H. {Van
  Ijzendoorn}, ``{Threat-related attentional bias in anxious and nonanxious
  individuals: A meta-analytic study},'' {\em Psychological Bulletin},
  vol.~133, no.~1, pp.~1--24, 2007.

\bibitem{Brosch2011}
T.~Brosch, G.~Pourtois, D.~Sander, and P.~Vuilleumier, ``{Additive effects of
  emotional, endogenous, and exogenous attention: Behavioral and
  electrophysiological evidence},'' {\em Neuropsychologia}, vol.~49, no.~7,
  pp.~1779--1787, 2011.

\bibitem{Dolcos2004}
F.~Dolcos, K.~S. LaBar, and R.~Cabeza, ``{Interaction between the amygdala and
  the medial temporal lobe memory system predicts better memory for emotional
  events},'' {\em Neuron}, vol.~42, no.~5, pp.~855--863, 2004.

\bibitem{Phelps2004}
E.~A. Phelps, ``{Human emotion and memory: Interactions of the amygdala and
  hippocampal complex},'' {\em Current Opinion in Neurobiology}, vol.~14,
  no.~2, pp.~198--202, 2004.

\bibitem{Sharot2004}
T.~Sharot, M.~R. Delgado, and E.~A. Phelps, ``{How emotion enhances the feeling
  of remembering},'' {\em Nature Neuroscience}, vol.~7, no.~12, pp.~1376--1380,
  2004.

\bibitem{Phelps2008}
E.~A. Phelps and T.~Sharot, ``{How (and why) emotion enhances the subjective
  sense of recollection},'' {\em Current Directions in Psychological Science},
  vol.~17, no.~2, pp.~147--152, 2008.

\bibitem{Phelps2006}
E.~A. Phelps, S.~Ling, and M.~Carrasco, ``{Emotion facilitates perception and
  potentiates the perceptual benefits of attention},'' {\em Psychological
  Science}, vol.~17, no.~4, pp.~292--299, 2006.

\bibitem{Brosch2010}
T.~Brosch, G.~Pourtois, and D.~Sander, ``{The perception and categorisation of
  emotional stimuli: A review},'' {\em Cognition and Emotion}, vol.~24, no.~3,
  pp.~377--400, 2010.

\bibitem{Spence1995}
S.~Spence, ``{Descartes' Error: Emotion, Reason and the Human Brain},'' {\em
  BMJ}, vol.~310, pp.~1213--1213, 1995.

\bibitem{Bechara1997}
A.~Bechara, H.~Damasio, D.~Tranel, and A.~R. Damasio, ``{Deciding
  advantageously before knowing the advantageous strategy},'' {\em Science},
  vol.~275, no.~5304, pp.~1293--1295, 1997.

\bibitem{Bechara2005}
A.~Bechara, H.~Damasio, D.~Tranel, and A.~R. Damasio, ``{The Iowa Gambling Task
  and the somatic marker hypothesis: Some questions and answers},'' {\em Trends
  in Cognitive Sciences}, vol.~9, no.~4, pp.~159--162, 2005.

\bibitem{Pantic2000}
M.~Pantic and L.~J.~M. Rothkrantz, ``{Automatic analysis of facial expressions:
  The state of the art},'' {\em IEEE Transactions on Pattern Analysis and
  Machine Intelligence}, vol.~22, no.~12, pp.~1424--1445, 2000.

\bibitem{Hanjalic2005}
A.~Hanjalic and L.~Q. Xu, ``{Affective video content representation and
  modeling},'' {\em IEEE Transactions on Multimedia}, vol.~7, no.~1,
  pp.~143--154, 2005.

\bibitem{ElKaliouby2005}
R.~{El Kaliouby} and P.~Robinson, ``{Real-time inference of complex mental
  states from facial expressions and head gestures},'' in {\em 2004 Conference
  on Computer Vision and Pattern Recognition Workshop}, pp.~154--154, 2004.

\bibitem{Yang2008}
Y.~H. Yang, Y.~C. Lin, Y.~F. Su, and H.~H. Chen, ``{A regression approach to
  music emotion recognition},'' {\em IEEE Transactions on Audio, Speech and
  Language Processing}, vol.~16, no.~2, pp.~448--457, 2008.

\bibitem{Zeng2009}
Z.~Zeng, M.~Pantic, G.~I. Roisman, and T.~S. Huang, ``{A survey of affect
  recognition methods: Audio, visual, and spontaneous expressions},'' {\em IEEE
  Transactions on Pattern Analysis and Machine Intelligence}, vol.~31, no.~1,
  pp.~39--58, 2009.

\bibitem{Schuller2010}
B.~Schuller, B.~Vlasenko, F.~Eyben, M.~W{\"{o}}llmer, A.~Stuhlsatz,
  A.~Wendemuth, and G.~Rigoll, ``{Cross-Corpus acoustic emotion recognition:
  Variances and strategies},'' {\em IEEE Transactions on Affective Computing},
  vol.~1, no.~2, pp.~119--131, 2010.

\bibitem{Polzehl2011}
T.~Polzehl, A.~Schmitt, F.~Metze, and M.~Wagner, ``{Anger recognition in speech
  using acoustic and linguistic cues},'' {\em Speech Communication}, vol.~53,
  no.~9-10, pp.~1198--1209, 2011.

\bibitem{Schuller2011}
B.~Schuller, A.~Batliner, S.~Steidl, and D.~Seppi, ``{Recognising realistic
  emotions and affect in speech: State of the art and lessons learnt from the
  first challenge},'' {\em Speech Communication}, vol.~53, no.~9-10,
  pp.~1062--1087, 2011.

\bibitem{Phillips2003}
M.~L. Phillips, W.~C. Drevets, S.~L. Rauch, and R.~Lane, ``{Neurobiology of
  emotion perception I: The neural basis of normal emotion perception.},'' {\em
  Biological psychiatry}, vol.~54, no.~5, pp.~504--14, 2003.

\bibitem{Dalgleish2004}
T.~Dalgleish, ``{The emotional brain.},'' {\em Nat Rev Neurosci.}, vol.~5,
  no.~July, pp.~252--256, 2004.

\bibitem{Redondo2014}
R.~L. Redondo, J.~Kim, A.~L. Arons, S.~Ramirez, X.~Liu, and S.~Tonegawa,
  ``{Bidirectional switch of the valence associated with a hippocampal
  contextual memory engram},'' {\em Nature}, vol.~513, 2014.

\bibitem{Ekman1983}
P.~Ekman, R.~W. Levenson, and W.~V. Friesen, ``{Autonomic nervous system
  activity distinguishes among emotions},'' {\em Science}, vol.~221, no.~4616,
  pp.~1208--1210, 1983.

\bibitem{Kreibig2010}
S.~D. Kreibig, ``{Autonomic nervous system activity in emotion: A review},''
  {\em Biological Psychology}, vol.~84, no.~3, pp.~394--421, 2010.

\bibitem{Ekman1987}
P.~Ekman, W.~V. Friesen, M.~O'Sullivan, A.~Chan, I.~Diacoyanni-Tarlatzis,
  K.~Heider, R.~Krause, W.~A. LeCompte, T.~Pitcairn, P.~E. Ricci-Bitti,
  K.~Scherer, M.~Tomita, and A.~Tzavaras, ``{Universals and Cultural
  Differences in the Judgments of Facial Expressions of Emotion},'' {\em
  Journal of Personality and Social Psychology}, vol.~53, no.~4, pp.~712--717,
  1987.

\bibitem{Scherer2001}
K.~R. Scherer, R.~Banse, and H.~G. Wallbott, ``{Emotion inferences from vocal
  expression correlate across languages and cultures},'' {\em Journal of
  Cross-Cultural Psychology}, vol.~32, no.~1, pp.~76--92, 2001.

\bibitem{Stahl2016}
S.~E. Stahl, H.-S. An, D.~M. Dinkel, J.~M. Noble, and J.-M. Lee, ``{How
  accurate are the wrist-based heart rate monitors during walking and running
  activities? Are they accurate enough?},'' {\em BMJ Open Sport {\&} Exercise
  Medicine}, vol.~2, 2016.

\bibitem{Wallen2016}
M.~P. Wallen, S.~R. Gomersall, S.~E. Keating, U.~Wisl{\o}ff, and J.~S. Coombes,
  ``{Accuracy of heart rate watches: Implications for weight management},''
  {\em PLoS ONE}, vol.~11, pp.~1--11, 2016.

\bibitem{Tison2018}
G.~H. Tison, J.~M. Sanchez, B.~Ballinger, A.~Singh, J.~E. Olgin, M.~J.
  Pletcher, E.~Vittinghoff, E.~S. Lee, S.~M. Fan, R.~A. Gladstone, C.~Mikell,
  N.~Sohoni, J.~Hsieh, and G.~M. Marcus, ``{Passive detection of atrial
  fibrillation using a commercially available smartwatch},'' {\em JAMA
  Cardiology}, vol.~3, no.~5, pp.~409--416, 2018.

\bibitem{Weber1992}
E.~J. Weber, P.~C. Molenaar, and M.~W. van~der Molen, ``{A Nonstationarity Test
  for the Spectral Analysis of Physiological Time Series with an Application to
  Respiratory Sinus Arrhythmia},'' {\em Psychophysiology}, vol.~29, no.~1,
  pp.~55--65, 1992.

\bibitem{Sunagawa1998}
K.~Sunagawa, T.~Kawada, and T.~Nakahara, ``{Dynamic nonlinear vago-sympathetic
  interaction in regulating heart rate},'' {\em Heart and Vessels}, vol.~13,
  no.~4, pp.~157--174, 1998.

\bibitem{Johnson2016}
A.~E. Johnson, M.~M. Ghassemi, S.~Nemati, K.~E. Niehaus, D.~Clifton, and G.~D.
  Clifford, ``{Machine Learning and Decision Support in Critical Care},'' {\em
  Proceedings of the IEEE}, vol.~104, no.~2, pp.~444--466, 2016.

\bibitem{Jerritta2011}
S.~Jerritta, M.~Murugappan, R.~Nagarajan, and K.~Wan, ``{Physiological signals
  based human emotion recognition: A review},'' in {\em Proceedings - 2011 IEEE
  7th International Colloquium on Signal Processing and Its Applications, CSPA
  2011}, pp.~410--415, 2011.

\bibitem{Kim2008b}
J.~Kim and E.~Andr{\'{e}}, ``{Emotion recognition based on physiological
  changes in music listening},'' {\em IEEE Transactions on Pattern Analysis and
  Machine Intelligence}, vol.~30, no.~12, pp.~2067--2083, 2008.

\bibitem{Alzoubi2012}
O.~Alzoubi, S.~K. D'Mello, and R.~A. Calvo, ``{Detecting naturalistic
  expressions of nonbasic affect using physiological signals},'' {\em IEEE
  Transactions on Affective Computing}, vol.~3, no.~3, pp.~298--310, 2012.

\bibitem{Goshvarpour2017}
A.~Goshvarpour, A.~Abbasi, and A.~Goshvarpour, ``{An accurate emotion
  recognition system using ECG and GSR signals and matching pursuit method},''
  {\em Biomedical Journal}, vol.~40, no.~6, pp.~355--368, 2017.

\bibitem{Miranda-Correa2017b}
J.~A. Miranda-Correa, M.~K. Abadi, N.~Sebe, and I.~Patras, ``{AMIGOS: A Dataset
  for Affect, Personality and Mood Research on Individuals and Groups},'' {\em
  IEEE Transactions on Affective Computing}, vol.~PP, 2017.

\bibitem{Subramanian2016}
R.~Subramanian, S.~Member, J.~{Wache Student Member}, M.~{Khomami Abadi},
  S.~Member, R.~L. Vieriu, S.~Winkler, and N.~Sebe, ``{ASCERTAIN: Emotion and
  Personality Recognition using Commercial Sensors},'' {\em IEEE Transactions
  on Affective Computing}, vol.~9, no.~2, pp.~147--160, 2018.

\bibitem{Agrafioti2012}
F.~Agrafioti, D.~Hatzinakos, and A.~K. Anderson, ``{ECG pattern analysis for
  emotion detection},'' {\em IEEE Transactions on Affective Computing}, vol.~3,
  no.~1, pp.~102--115, 2012.

\bibitem{Katsigiannis2018}
S.~Katsigiannis and N.~Ramzan, ``{DREAMER: A Database for Emotion Recognition
  Through EEG and ECG Signals from Wireless Low-cost Off-the-Shelf Devices},''
  {\em IEEE Journal of Biomedical and Health Informatics}, vol.~22, no.~1,
  pp.~98--107, 2018.

\bibitem{Guo2016a}
H.~W. Guo, Y.~S. Huang, C.~H. Lin, J.~C. Chien, K.~Haraikawa, and J.~S. Shieh,
  ``{Heart Rate Variability Signal Features for Emotion Recognition by Using
  Principal Component Analysis and Support Vectors Machine},'' in {\em
  Proceedings - 2016 IEEE 16th International Conference on Bioinformatics and
  Bioengineering, BIBE 2016}, pp.~274--277, 2016.

\bibitem{Valenza2014b}
G.~Valenza, L.~Citi, A.~Lanat{\'{a}}, E.~P. Scilingo, and R.~Barbieri,
  ``{Revealing real-time emotional responses: A personalized assessment based
  on heartbeat dynamics},'' {\em Scientific Reports}, vol.~4, pp.~1--13, 2014.

\bibitem{Torres-Valencia2015}
C.~A. Torres-Valencia, H.~F. Garcia-Arias, M.~A. Lopez, and A.~A.
  Orozco-Gutierrez, ``{Comparative analysis of physiological signals and
  electroencephalogram (EEG) for multimodal emotion recognition using
  generative models},'' in {\em 2014 19th Symposium on Image, Signal Processing
  and Artificial Vision, STSIVA}, pp.~1--5, 2014.

\bibitem{Garcia2016}
H.~F. Garc{\'{i}}a, M.~A. {\'{A}}lvarez, and {\'{A}}.~A. Orozco, ``{Gaussian
  process dynamical models for multimodal affect recognition},'' in {\em
  Proceedings of the Annual International Conference of the IEEE Engineering in
  Medicine and Biology Society, EMBS}, pp.~850--853, IEEE, 2016.

\bibitem{Soleymani2012}
M.~Soleymani, J.~Lichtenauer, T.~Pun, and M.~Pantic, ``{A multimodal database
  for affect recognition and implicit tagging},'' {\em IEEE Transactions on
  Affective Computing}, vol.~3, no.~1, pp.~42--55, 2012.

\bibitem{Alhagry2017}
S.~Alhagry, A.~Aly, and R.~A., ``{Emotion Recognition based on EEG using LSTM
  Recurrent Neural Network},'' {\em International Journal of Advanced Computer
  Science and Applications}, vol.~8, no.~10, 2017.

\bibitem{Keren2017}
G.~Keren, T.~Kirschstein, E.~Marchi, F.~Ringeval, and B.~Schuller,
  ``{End-to-end learning for dimensional emotion recognition from physiological
  signals},'' in {\em Proceedings - IEEE International Conference on Multimedia
  and Expo}, pp.~985--990, 2017.

\bibitem{Barrett1998}
L.~F. Barrett, ``{Discrete Emotions or Dimensions? The Role of Valence Focus
  and Arousal Focus},'' {\em Cognition and Emotion}, 1998.

\bibitem{Kolodyazhniy2011}
V.~Kolodyazhniy, S.~D. Kreibig, J.~J. Gross, W.~T. Roth, and F.~H. Wilhelm,
  ``{An affective computing approach to physiological emotion specificity:
  Toward subject-independent and stimulus-independent classification of
  film-induced emotions},'' {\em Psychophysiology}, vol.~48, no.~7,
  pp.~908--922, 2011.

\bibitem{Ferdinando2016}
H.~Ferdinando, T.~Seppanen, and E.~Alasaarela, ``{Comparing features from ECG
  pattern and HRV analysis for emotion recognition system},'' in {\em CIBCB
  2016 - Annual IEEE International Conference on Computational Intelligence in
  Bioinformatics and Computational Biology}, pp.~1--6, 2016.

\bibitem{Ghahramani2015}
Z.~Ghahramani, ``{Probabilistic machine learning and artificial
  intelligence},'' {\em Nature}, vol.~521, no.~7553, pp.~452--459, 2015.

\bibitem{Neal1996}
R.~M. Neal, {\em {Bayesian Learning for Neural Networks}}.
\newblock 1996.

\bibitem{MacKay1992}
D.~J. MacKay, {\em {Bayesian Methods for Adaptive Models}}.
\newblock PhD thesis, California Institute of Technology, 1992.

\bibitem{Welling2011}
M.~Welling and Y.-W. Teh, ``{Bayesian learning via stochastic gradient Langevin
  dynamics},'' in {\em ICML Proceedings of the 28th International Conference on
  International Conference on Machine Learning}, pp.~681--688, 2011.

\bibitem{Hasenclever2015}
L.~Hasenclever, S.~Webb, T.~Lienart, S.~Vollmer, B.~Lakshminarayanan,
  C.~Blundell, and Y.~W. Teh, ``{Distributed Bayesian Learning with Stochastic
  Natural-gradient Expectation Propagation and the Posterior Server},'' {\em
  Journal of Machine Learning Research}, vol.~18, no.~1, pp.~3744--3780, 2017.

\bibitem{Graves2012}
A.~Graves, ``{Practical Variational Inference for Neural Networks},'' in {\em
  Proceedings of the 24th International Conference on Neural Information
  Processing Systems}, pp.~2348--2356, 2011.

\bibitem{Gal2016}
Y.~Gal and Z.~Ghahramani, ``{Dropout as a Bayesian Approximation: Representing
  Model Uncertainty in Deep Learning},'' in {\em Proceedings of the 33rd
  International Conference on Machine Learning (ICML-16)}, pp.~1050--1059,
  2016.

\bibitem{Sutskever2014}
I.~Sutskever, G.~Hinton, A.~Krizhevsky, and R.~R. Salakhutdinov, ``{Dropout : A
  Simple Way to Prevent Neural Networks from Overfitting},'' {\em Journal of
  Machine Learning Research}, vol.~15, pp.~1929--1958, 2014.

\bibitem{Morris1995}
J.~D. Morris, ``{SAM: The Self-Assessment Manikin - An efficient cross-cultural
  measurement of emotional response},'' {\em Journal of Advertising Research},
  vol.~35, no.~6, pp.~63--68, 1995.

\bibitem{Gabert-Quillen2015}
C.~A. Gabert-Quillen, E.~E. Bartolini, B.~T. Abravanel, and C.~A. Sanislow,
  ``{Ratings for emotion film clips},'' {\em Behavior Research Methods},
  vol.~47, no.~3, pp.~773--787, 2015.

\bibitem{Christov2004}
I.~I. Christov, ``{Real time electrocardiogram QRS detection using combined
  adaptive threshold},'' {\em BioMedical Engineering Online}, vol.~3, no.~1,
  p.~28, 2004.

\bibitem{He2015}
K.~He, X.~Zhang, S.~Ren, and J.~Sun, ``{Delving deep into rectifiers:
  Surpassing human-level performance on imagenet classification},'' in {\em
  Proceedings of the IEEE International Conference on Computer Vision}, 2015.

\bibitem{Kingma2014}
P.~D. Kingma and J.~Ba, ``{Adam: A Method for Stochastic Optimization},'' {\em
  International Conference on Learning Representations}, 2014.

\bibitem{GoogleResearch2015}
GoogleResearch, ``{TensorFlow: Large-scale machine learning on heterogeneous
  systems},'' {\em Google Research}, 2015.

\bibitem{Pietila2017}
J.~Pietil{\"{a}}, S.~Mehrang, J.~Tolonen, E.~Helander, H.~Jimison, M.~Pavel,
  and I.~Korhonen, ``{Evaluation of the accuracy and reliability for
  photoplethysmography based heart rate and beat-to-beat detection during daily
  activities},'' in {\em IFMBE Proceedings}, 2017.

\bibitem{Harju2018}
J.~Harju, A.~Tarniceriu, J.~Parak, A.~Vehkaoja, A.~Yli-Hankala, and
  I.~Korhonen, ``{Monitoring of heart rate and inter-beat intervals with wrist
  plethysmography in patients with atrial fibrillation},'' {\em Physiological
  Measurement}, vol.~39, no.~6, 2018.

\end{thebibliography}

\begin{IEEEbiography}[{\includegraphics[width=1in,height=1.25in,clip,keepaspectratio]{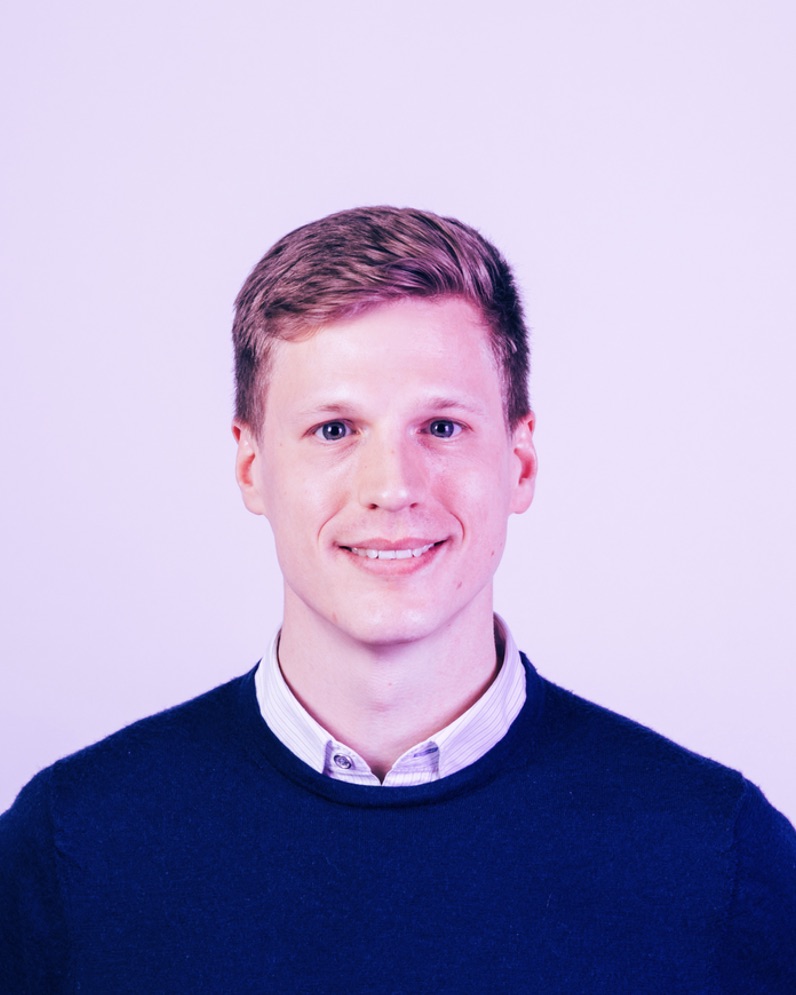}}]{Ross Harper}
	completed a PhD in computational neuroscience and an MRes in mathematical modelling at University College London. Prior to this, Ross studied natural sciences at Cambridge University. Ross is the CEO of Limbic Ltd, a company developing emotion recognition technologies. Ross is the recipient of the Lloyd’s Bank New Entrepreneur of the Year (2018) award. His current research interests include, affective computing, probabilistic modelling of biometric time series data, and wearable computing. 
\end{IEEEbiography}

\begin{IEEEbiography}[{\includegraphics[width=1in,height=1.25in,clip,keepaspectratio]{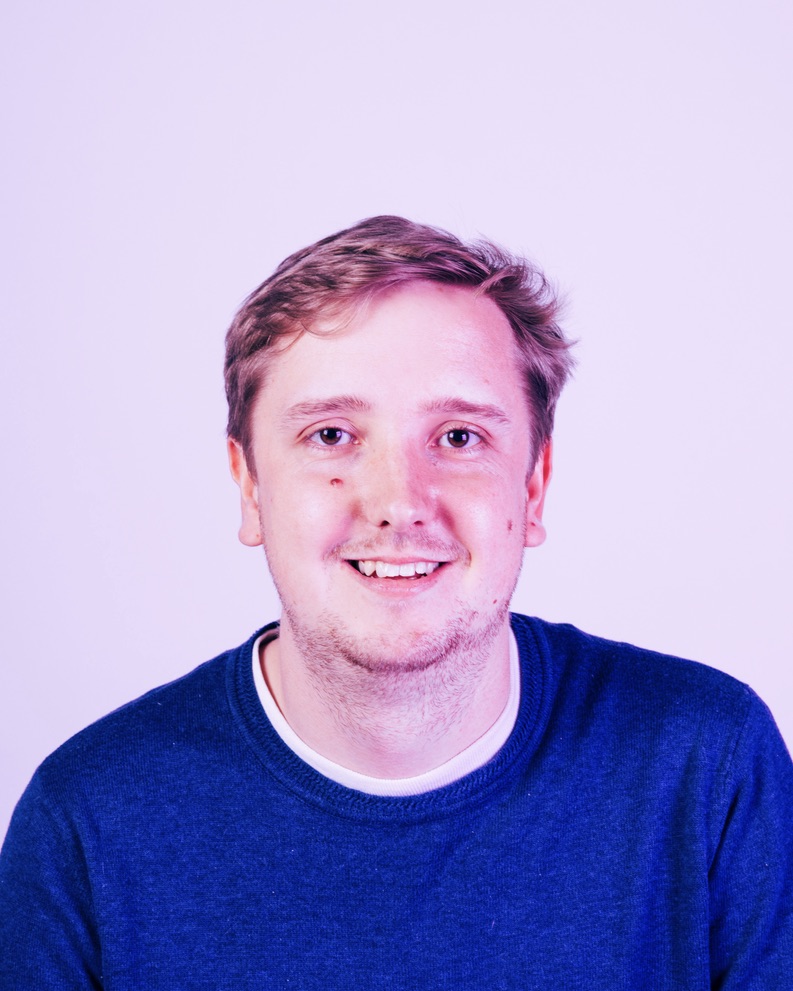}}]{Joshua Southern}
obtained an Msc in applied mathematics from Imperial College London, and a Bsc in physics from Bath University. He is a machine learning researcher at Limbic Ltd, and previously completed an internship at IBM. His research interests include deep learning, time series analysis, and complex networks. 
\end{IEEEbiography}

\end{document}